\documentclass{article}
\usepackage{ijcai19}

\usepackage{times}
\usepackage{soul}
\usepackage{url}
\usepackage[hidelinks]{hyperref}
\usepackage[utf8]{inputenc}
\usepackage[small]{caption}
\usepackage{graphicx}
\usepackage{amsmath}
\usepackage{booktabs}
\usepackage{multirow}
\usepackage{algorithm}
\usepackage{algorithmic}

\usepackage{amssymb}
\usepackage{bbding}
\title{Combined Dynamic Virtual Spatiotemporal Graph Mapping for Traffic Prediction}
\author{
    Yingming Pu
    \affiliations
    Westlake University, China \emails
    puyingming@westlake.edu.cn
}

\begin{document}

\maketitle

\begin{abstract}

The continuous expansion of the urban construction scale has recently contributed to the demand for the dynamics of traffic intersections that are managed, making adaptive modellings become a hot topic. Existing deep learning methods are powerful to fit complex heterogeneous graphs. However, they still have drawbacks, which can be roughly classified into two categories, 1) spatiotemporal async-modelling approaches separately consider temporal and spatial dependencies, resulting in weak generalization and large instability while aggregating; 2) spatiotemporal sync-modelling is hard to capture long-term temporal dependencies because of the local receptive field. In order to overcome above challenges, a \textbf{C}ombined \textbf{D}ynamic \textbf{V}irtual spatiotemporal \textbf{G}raph \textbf{M}apping \textbf{(CDVGM)} is proposed in this work. The contributions are the following: 1) a dynamic virtual graph Laplacian ($DVGL$) is designed, which considers both the spatial signal passing and the temporal features simultaneously; 2) the Long-term Temporal Strengthen model ($LT^2S$) for improving the stability of time series forecasting; Extensive experiments demonstrate that CDVGM has excellent performances of fast convergence speed and low resource consumption and achieves the current SOTA effect in terms of both accuracy and generalization. The code is available at \hyperlink{https://github.com/Dandelionym/CDVGM.}{https://github.com/Dandelionym/CDVGM.}
\end{abstract}

\section{Introduction}
Traffic flow forecasting is a fundamental intelligent transportation task that requires predicting the traffic flow in all interactions of a road map \cite{FOGS,IM1,IM2,IM3,IM4,IM5}. Traffic flow forecasting subsumes a series prediction task, as the flow of a road network can be thought of as a set of nodes’ temporal series prediction task.  Therefore, it is a more challenging task than time series prediction by requiring both spatial-level modelling and temporal-level prediction because of the complex dependencies of spatial dimensions.

\begin{figure}
  \begin{center}
  \includegraphics[width=3.2in]{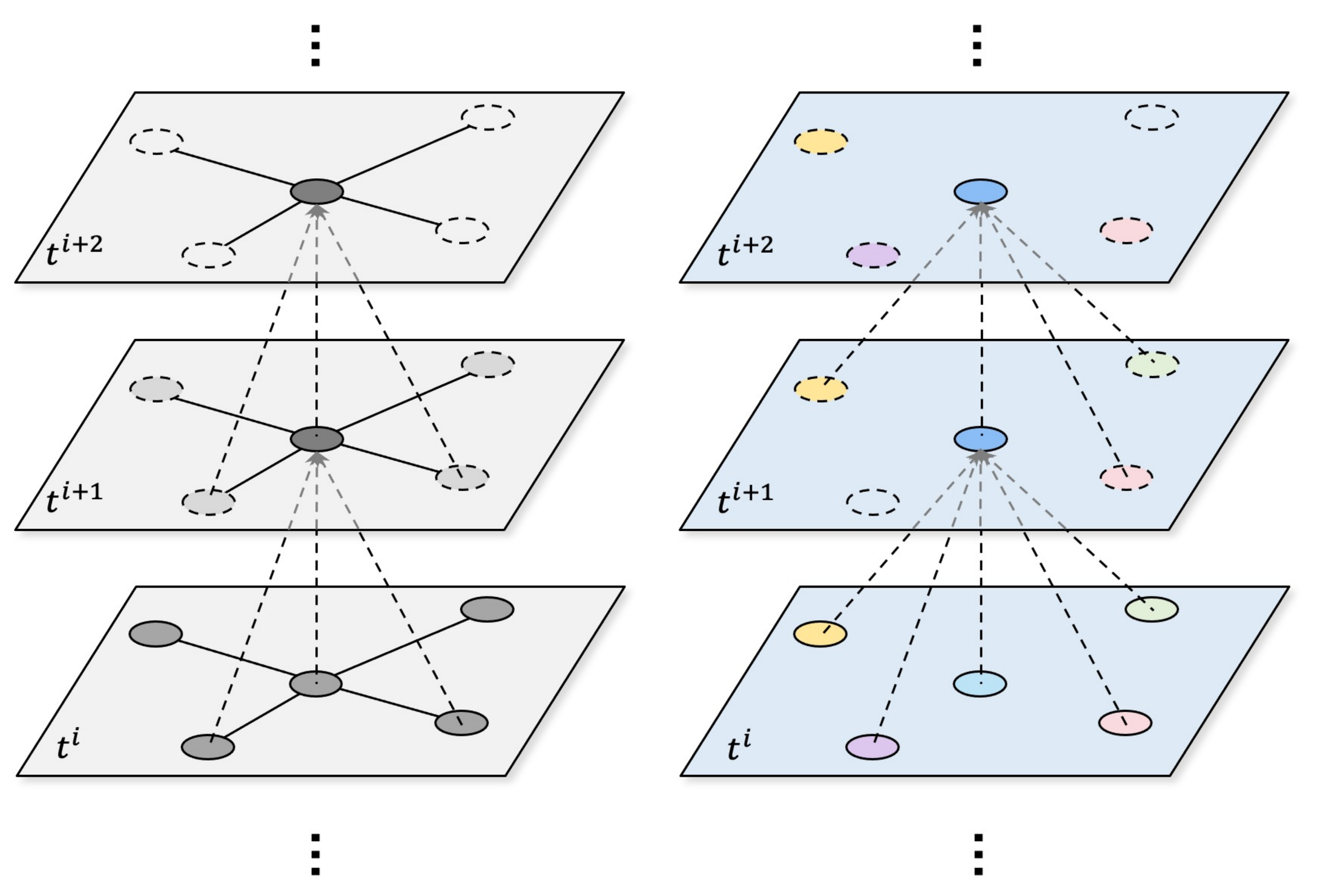}
  \caption{General topology-based methods utilize a fixed road map to represent spatial connection or dynamics aggregated with the road map as spatial connection(left), which is built on the assumption that there is no indirect societal level of connection; Our dynamic virtual spatiotemporal graph (right), has the ability to model low-high-dimension-oriented dependencies without topological information adaptively.}
  \end{center}
\end{figure}

Recently, significant progress has been made to address the topology-based traffic prediction task \cite{TM1,TM2}. However, the topological structure of a graph can be dynamically changing in the real world and the traffic crash can lead to unavailable crossroads. Latest methods such as DGCN \cite{DGCN}, DCRNN \cite{DCRNN}, STFGNN \cite{STFGNN} and S2TAT \cite{S2TAT} have significantly improved the fitness of spatiotemporal modelling. However, on the one hand, their competitive result still relies on topological connections that are not adaptive enough due to traffic jams or crashes, as shown in Fig 1. On the other hand, it is generally confused whether it is better to use the space-time asynchronous method or the space-time synchronization method because both of them have their advantages, for example, a synchronous method such as ASTGCN \cite{ASTGCN} and DCRNN \cite{DCRNN} often have the weakness of local receptive field, which lead to the shortage of modelling long-term dependencies and asynchronous methods such as STFGNN \cite{STFGNN} and S2TAT \cite{S2TAT} lack the ability of dependency’s representation because the spatiotemporal graph structure inherently has the complex coupling in both spatial and temporal dimensions. To our knowledge, existing methods hardly take into account each of the important conditions mentioned above simultaneously because of the consideration of computational complexity or inference speed or even modelling ability. 

To overcome these shortages, the Combined Dynamic Virtual spatiotemporal Graph Mapping (CDVGM) is proposed. In this work, we explore capturing dynamic correlations between spatial and temporal dimensions without any topological additions and the stability of the prediction. The work draws on cross-entropy theory and uses it as the basis for the asymmetric study of node correlations in the construction of dynamic Laplacian. In this work, all traffic nodes are located in the same quantifying space to compute the differences by cross-entropy of history flow data. We treat the result as the expression of correlations of each two nodes in the case of incorporating temporal features. Therefore, the direction is also addressed as there are up-streams and down-streams objectively in the road of the real. Unlike existing methods, CDVGM efficiently generates adaptive Laplacian through history flows in a series of states even if the road network is changing, e.g. some interactions might be unavailable due to the crash. Finally, we find that the graph Laplacian operators with different order of magnitude scales can better represent the key nodes in the spatial network and find the high-energy regions existing in the corresponding road network, which is more conducive to the transmission of induced graph signals. Besides, the stability of the mid-range forecasting problem also got relief by the way of $LT^2S$ module, which takes the trend as the key of the prediction task and improves the whole accuracy even in complex couplings of spatiotemporal dependencies.
\begin{figure*}[htbp]
\begin{center}
\includegraphics[width=0.8\textwidth,height=0.42\textwidth]{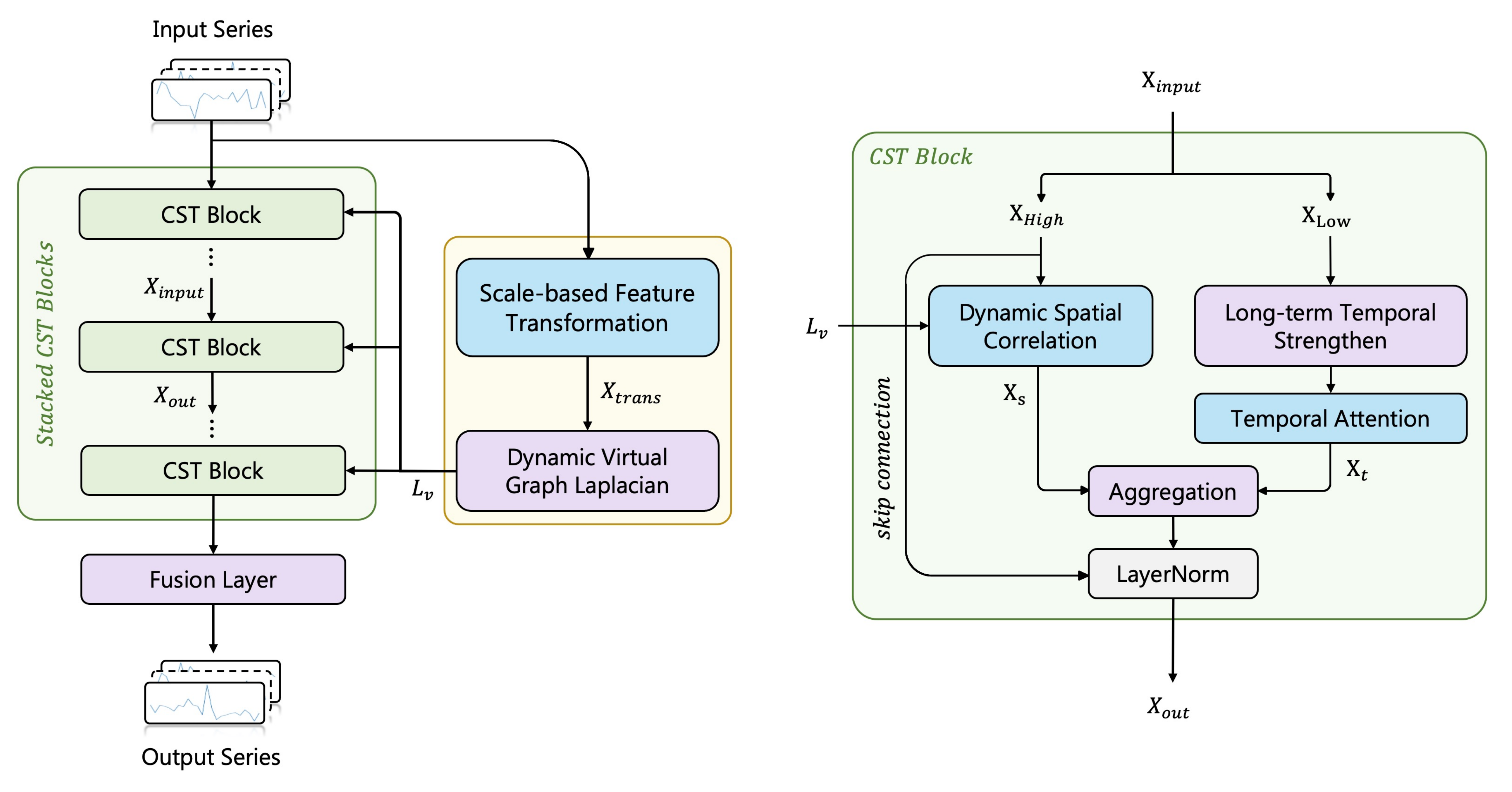}
\end{center}
\caption{The architecture of CDVGM framework (left) and details of CST Block (right). The framework takes in a series of inputs and passes them into both stacked CST Blocks and Dynamic Virtual Graph Laplacian for modelling. Several green background blocks mainly capture the spatiotemporal dependencies while the yellow background block is for dynamic Laplacian construction in our framework.}
\end{figure*}
\begin{enumerate}
    \item[1)]{The first topological-structure-free framework with dynamic dependencies modelling is proposed, which combined both synchronous and asynchronous advantages within fast convergence speed and excellent prediction accuracy.}
    \item[2)]{The Laplacian is carefully designed by a dynamic virtual graph for graph signal passing, which considers the temporal correlations and spatial connections simultaneously in a time-series-based way by similarity theory.}
    \item[3)]{A Long-Term Temporal Strengthen ($LT^2S$) module is proposed to enhance the perception of long-range dependencies with flexibility. It gives a simple but effective way to the series prediction task.}
    \item[4)]{Extensive experiments on four benchmarks demonstrate that the proposed framework outperforms many recent state-of-the-art methods, implying that CDVGM has the best predictive ability and application value so far.}
\end{enumerate}

\section{Related works}
\subsection{Traffic Prediction}
Traffic forecasting has many application values in smart city construction. Traffic data is often viewed as a spatiotemporal graph. Due to the limited modelling ability of the early statistical methods \cite{ARIMA}, they only consider the temporal dimension for modelling, ignoring the geographic effect of the spatial dimension, which leads to the fundamental defects of such methods. Subsequently, STGCN \cite{STGCN} and DCRNN \cite{DCRNN} model the spatial dimension through a deep learning method with GLU or RNN for temporal prediction asynchronously while methods such as STSGCN \cite{STSGCN} and STFGNN \cite{STFGNN} use a local graph to represent spatiotemporal correlation synchronously. S2TAT \cite{S2TAT} uses a time-oriented graph convolution network to improve the ability of spatiotemporal perception. Owing to the challenges of spatiotemporal modelling, attention mechanism is adopted such as ASTGCN \cite{ASTGCN} and DGCN \cite{DGCN} etc. Except this, STGODE \cite{STGODE} uses spatial-based adjacency matrices and semantic-based adjacency matrices to reflect spatial dependencies, and perceive high-dimensional spatiotemporal correlations through ordinary differential equations. ST-3DNet \cite{ST-3DNet} first utilizes a 3D-convolution operator for spatiotemporal graph modelling and ST-ResNet \cite{ST-ResNet} considers the fact of temporal characteristics of crowd movement to fit the real-world situation better.

In a word, existing methods normally either consider the complex coupling as a separate modelling problem wrongly within the IID assumption or ignore the direct or indirect entanglement that exists between potential spatiotemporal dependencies, which leads to the drawbacks of low generalizability. Unlike prior works, our novelty is that proposing a topological-free virtual graph designed by combining spatiotemporal dimensions and equipped with a temporal strengthen strategy asynchronously to boost performance for the prediction task.
\subsection{Graph convolutional neural network}
The graph convolutional neural network realizes the operation of convolution of non-Euclidean graph data, and GNN has had a significant impact in the fields of social relationship mining and chemical biology. It can be roughly divided into two categories, one is a spectral graph neural network based on spectral graph theory, and the other is a graph network based on the spatial method. Among them, ChebyNet \cite{Chebynet} uses Chebyshev polynomials to approximate the Laplacian operator of spectral graph decomposition, which greatly reduces the computational complexity and is a typical representative of spectral graph neural networks. Then graph convolution neural network (GCN) \cite{GCN} simplifies ChebyNet with a first-order polynomial, which removes the hyperparameter of K-level adjacent and becomes the cornerstone of spatial graph neural network. Space-based GCN generalizes convolution in Euclidean space to work on graph data. For example, GraphSAGE \cite{GraphSage} transmits the neighbour node's signal through an adjacency matrix before aggregating features, and Graph Attention Network (GAT) \cite{GAT} weight node signals by attention mechanism. Simplified Graph Convolution Network (SGC) \cite{SGC} removes the non-linear activations in   hidden layers for local averaging of feature propagation. In general, GCN has a strong ability to perceive spatial dependencies in graph data, and for this reason, the GCN is crucial to extract spatial dependencies in most methods.
\subsection{Synchronous and Asynchronous Spatiotemporal Modelling}
Spatiotemporal modelling is an open challenge in many research fields, such as medical analysis and social network mining \cite{STref1,STref2,STref3,STref4,STref5}.  Owing to the complex coupling relationships in spatial and temporal dimensions, existing methods can be divided into two parts:
\begin{enumerate}
    \item [1)]The synchronous methods, which aim to simultaneously perceive spatiotemporal dependencies, believe that time and space are locally entangled and driven by the action of nodes. In this way, STSGCN \cite{STSGCN} designs a local graph to model localized dependencies, and STFGNN \cite{STFGNN} considers both spatial and temporal behaviours synchronously through a spatiotemporal fusion graph in a data-driven manner. $S^2TAT$ \cite{S2TAT} uses a time-based graph convolution operation to express temporal dependencies while spatial modelling. Auto-DSTSG \cite{AutoDSTSG}  builds an automated dilated spatiotemporal synchronous graph loaded with neural network architecture search to capture long- and short-term spatiotemporal dependencies. MSGAT \cite{MSGAT} learns three embeddings to respectively but synchronously represent the traffic data-based channel, temporal, and spatial relations between nodes by specific graph attention designs while the ESTNet \cite{ESTNET} utilizes a 3D convolution unit for spatiotemporal modelling task. 
    \item [2)]The asynchronous methods, which separately focus on spatial and temporal dependencies with unfixed order are widely used because of the flexibility in feature extraction. STGCN \cite{STGCN} is the first framework of asynchronous modelling through the stacking of the ST-Conv Blocks. DCRNN \cite{DCRNN} uses Diffusion convolution and RNN for spatial and temporal respectively. Except for this, ASTGCN \cite{ASTGCN} adopts an attention mechanism to capture latent spatiotemporal correlations. And DGCN \cite{DGCN}, Dynamic Graph \cite{DynamicGraph}, STSSN \cite{STSSN} also utilize graph convolution network and time series modelling methods to address the difficulties. In addition, asynchronous methods always need an aggregation module to fusion the spatial and temporal dependencies.
\end{enumerate}
  
Both of these two methods have their advantages and disadvantages. The former is hard to capture global correlations because of the receptive field and the latter always deeply relies on the way of aggregations. Therefore, regarding the modelling process of the spatiotemporal graphs, both synchronous and asynchronous are needed to gather superiority of better capture latent dependencies among traffic data.

\section{Methodology}

\subsection{Preliminaries}

Define spatial graph  G=$\langle$V, E$\rangle $, $\left | V \right |$ = N, where V and E are collections of nodes and edges respectively. Adjacency matrix marked as $A \in\mathbb{R}^{N\times N}$ is generated by graph G. Node features $X \in \mathbb{R}^{F\times N\times T}$, where $F$ is the dimension of features.

The goal of the traffic flow prediction task is to construct a mapping $\ \mathcal{F}\left(\cdot\right)$ from historical data to future data:
$$\left[X^{t-T+1}, \ldots, X^{t}\right] \stackrel{\mathcal{F}(\cdot)}{\longrightarrow}\left[X^{t+1}, \ldots, X^{t+T^{\prime}}\right]$$
where T represents the length of the historical sequence, $T^{'}$ is the length of the prediction sequence. 

The proposed framework is shown in Fig.2. It contains 1) cascaded CST Blocks for perceiving high-dimensional spatiotemporal correlations; 2) a Scale-based Feature Transformation module for feature fusion and normalization; 3) a Dynamic Virtual Graph Laplacian module for building dynamic virtual graphs based on node features. and 4) Fusion Layer for temporal prediction. The processes of algorithms are listed in \textbf{Algorithm 1}, and each sub-module of CDVGM is explained one by one below.

\subsection{Scale-based Feature Transformation}
Scale-based feature transformation is the first sub-module of CDVGM for forwarding propagation, its input $X_{in}\in\mathbb{R}^{F\times N\times T}\ $, where F is the dimension of features and the input is without any normalization. This module is formulated as:
\begin{equation}
X_{trans}=\sigma\left({Conv}_{1\times1}\left(X_{input}\right)\right)
\end{equation}
where $\sigma$ is an activation function, such as sigmoid. The meaning of feature transformation is to fuse multiple features into one dimension. On the one hand, it is to project high-dimensional features into a low-dimensional vector space through linear transformation to neutralize the meaning of each item in the original feature of nodes. On the other hand, it is easier for the model to absorb effective features by the first aggregation and then calculating. Thus, we get the output $X_{trans}\in\mathbb{R}^{N\times T}\ $ for constructing dynamic Laplacian.

% \begin{algorithm}[tb]
% \caption{Example algorithm}
% \label{alg:algorithm}
% \textbf{Input}: Your algorithm's input\\
% \textbf{Parameter}: Optional list of parameters\\
% \textbf{Output}: Your algorithm's output
% \begin{algorithmic}[1] %[1] enables line numbers
% \STATE Let $t=0$.
% \WHILE{condition}
% \STATE Do some action.
% \IF {conditional}
% \STATE Perform task A.
% \ELSE
% \STATE Perform task B.
% \ENDIF
% \ENDWHILE
% \STATE \textbf{return} solution
% \end{algorithmic}
% \end{algorithm}

\begin{algorithm}[tb]
    \caption{Algorithm in CDVGM}
    \textbf{Input}: $X_{input}$\\
    \textbf{Parameter}: $K$ for ChebyNet\\
    \textbf{Output}: $X_{out}$
    \begin{algorithmic}[1]
        \STATE \textbf{\{Scale-based Feature Transformation\}} \\
        \STATE ${X_{in} \gets \sigma(Conv_{1*1}(X_{input}))}$ \\ 
        \STATE \textbf{\{Dynamic Virtual Graph Laplacian\}} \\
        \STATE ${L_{t} \gets P_{h}P_{h}^{T} + P_{b}}$ \\
        \STATE ${L_{c} \gets -X_{in} \otimes \log{(1+X_{in}^{T})}}$ \\
        \STATE ${L_{v} \gets \theta \cdot \sigma (L_{t} + L_{c})}$ \\
        \STATE \textbf{\{Several CST Blocks\}}
        \FOR{$l=1, 2, ..., N_{block}$}{
            \IF{$l < N$} 
            \STATE {${L_{v} \gets Mean(L_{v}) - (L_{v} \otimes L_{v}^T)}$}\\
            \ENDIF
            \STATE ${X_{High}, X_{Low} \gets X_{in}}$ \\
            \STATE ${X_s \gets ChebyNet(L_{v}, X_{High}, K)}$ \\
            \STATE ${\textbf{\{Long-term Temporal Strengthen\}}}$\\
            \STATE ${X_{conv} \gets \sigma(Conv(X_{Low}))}$ \\ 
            \STATE ${X_t \gets Concat(X_{Low}^0, X_{conv}, X_{Low}^{-1})}$ \\
            \STATE ${A_x \gets SelfAttention(X_t)}$ \\
            \STATE  \textbf{\{Aggregation\}} \\
            \STATE ${X_{st} \gets Conv(A_x X_s)}$ \\
            \STATE ${X_{st} \gets LayerNorm(X_{High} + X_{st})}$ \\
        }
        \ENDFOR
        \STATE \textbf{\{Fusion Layer\}} \\ 
        \STATE $X_{head} = Conv(X_{st}[:T:])$ \\
        \STATE $X_{tail} = Conv(X_{st}[T::])$ \\ 
        \STATE $X_{out} = TCN(X_{head} + X_{tail})$
        \STATE \textbf{return} $X_{out}$
    \end{algorithmic}
\end{algorithm}

\subsection{Dynamic Virtual Graph Laplacian}
In general, it is a well-established fact in practical spatiotemporal graph modelling that the node scale is usually much larger than the temporal series scale. Graph convolutional neural network is awesome because it has the ability of computing non-Euclidean data such as graphs. Although many variants of GCN have been proposed, they still lack flexibility. As for the spectral domain method and the spatial method, the former has a stronger graph feature propagation ability than the spatial method if the graph Laplacian operator with high-dimensional spatial correlation is established, so the ChebyNet \cite{Chebynet} is used to extract spatial dependencies in CDVGM. The ChebyNet adopts Chebyshev polynomials to approximate the graph Laplacian to reduce the prior computational complexity defined by spectral decomposition:
\begin{equation}
\begin{aligned}
x\ \ast_\mathcal{G}y&=\sigma\left(U\left(\sum_{k=0}^{K-1}{\theta_k\Lambda^k}\right)U^Tx\right) \\
&=\ \sigma\left(\left(\theta_0I+\theta_1L+\ldots+\theta_{K-1}L^{K-1}\right)x\right)
\end{aligned}
\end{equation}
where L is Laplacian matrix, $\theta_0\in\left[0,\ K-1\right]$ is polynomial parameters. 

The Chebyshev network approximates spectral decomposition by information discarding, which greatly reduces the computational complexity and omits the aggregation of higher-order neighbour signals. It is not difficult to find that if the signal aggregation order of a graph node is too high and each node aggregates too many neighbour nodes, the signal of the node will expose, which inevitably leads to excessive smoothing of the graph. From this aspect, the dynamic evolution of graphs relies heavily on the layer awareness of input features as high dimensional features often have more couplings that are hard to be controlled. 

Therefore, a high-order trend matrix and a low-order connectivity matrix are inferred and integrated together without entanglement to obtain a unique representation of the signal of a spatiotemporal graph.

In this work, the trend matrix through self-learning and the connectivity matrix based on the cross-entropy theory are designed and formulated in Equations (3) and (4), to extract the network trend information in the time dimension from the latent space. 
% Finally, the two are linearly superimposed, and the scale factor is used to control the feature scale of different graphs.

\begin{equation}
\begin{aligned}
L_{t}&=P_hP_h^T+P_b
\end{aligned}
\end{equation}
\begin{equation}
\begin{aligned}
L_{c}&={-X}_{in} \otimes log\ {\left(1+X_{in}^T\right)}
\end{aligned}
\end{equation}
\begin{equation}
L_{v}=\theta \cdot \sigma(L_{t}+L_{c})
\end{equation}
where $P_h$, ${P_b}\ \in\ \mathbb{R}^{N\times N}$ are learnable parameters, $\sigma_1(\cdot)$, $\sigma_2(\cdot)$ are activation functions, such as leakyReLU and $\theta$ is a hyperparameter for scaling the order of magnitude of the dynamic adjacency matrix and $ \otimes $ is matrix product.

The DVGL module absorbs the cross-entropy theory and formally constructs the node similarity expression that can be calculated in parallel through matrix operations, so the computation efficiency is excellent. $L_{t}$ represents the weak spatial trends between different nodes through a learning method, and $L_{c}$ uses cross-entropy to represent the main active nodes in the road network and their associations with other nodes using temporal series similarity. After combining the two, the generalized dynamic virtual graph Laplacian $L_{v}\in\mathbb{R}^{N\times N}$ that can represent both nodes similarity from spatial dimension and node feature series from temporal dimension is given, which is adopted to represent the dependencies of the road network from the perspective of both temporal and spatial level.
%
%
% =======
% FIG. 03
% =======
\begin{figure}
  \begin{center}
  \includegraphics[width=3.2in]{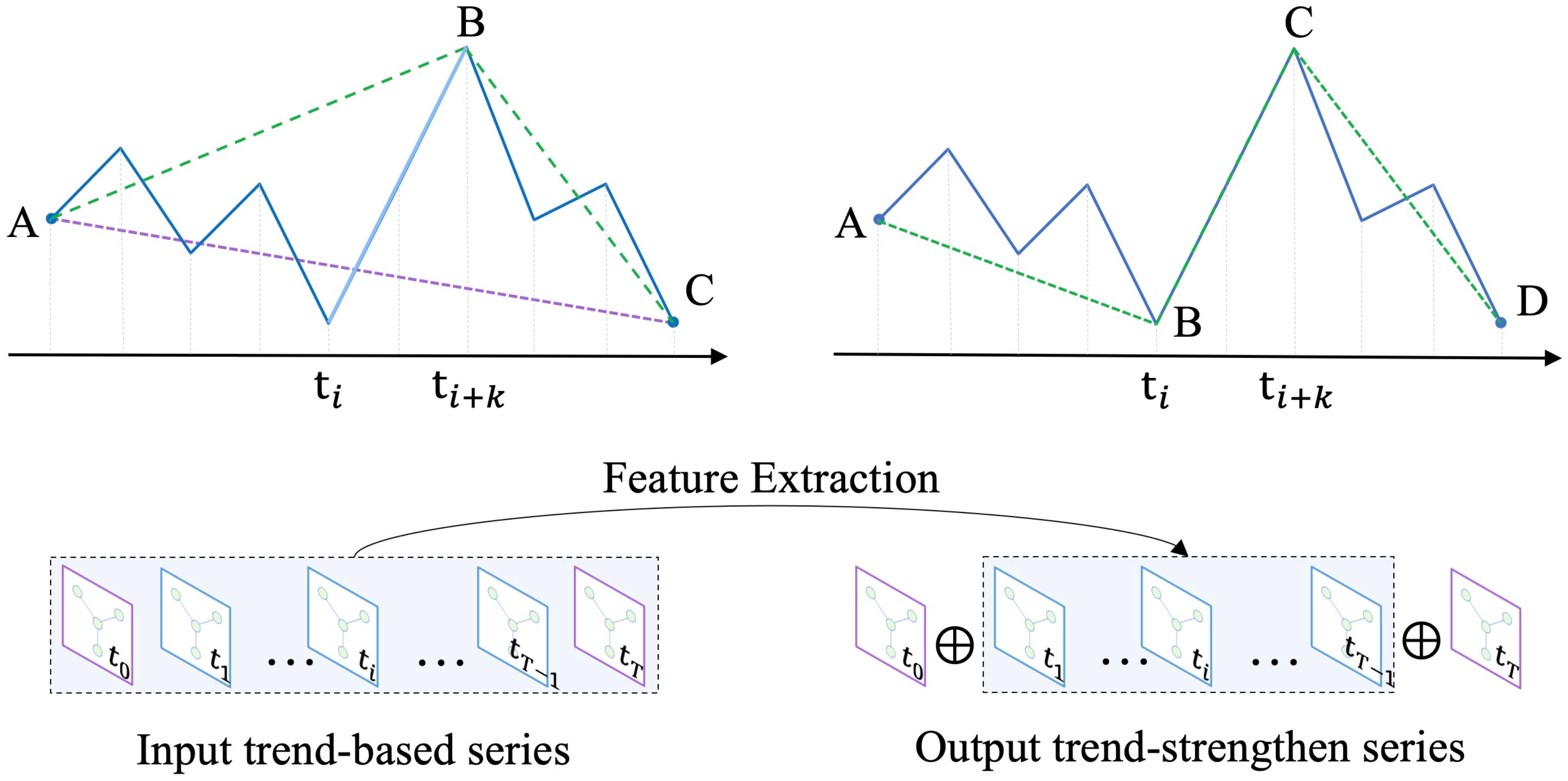}
  \caption{Long-term temporal strengthen module. It is a fact that a long sequence has a local trend in different stages (i.e. AB and BC), which is like the green dashed line drawn in the top two examples. Moreover, the sequence has a global trend as drawn with a purple dash line in the top left example (i.e. AC), which informs that the global trend can be captured by multi-local trends with some methods, as shown in the top right (i.e. AB, BC and CD).}
  \end{center}
\end{figure}
\subsection{Long-term Temporal Strengthen}
In the task of series prediction, long-range dependency modelling is always a challenging task because of the uncertainty of the real. Existing methods mainly rely on the outstanding performance of the attention mechanism. However, a large number of data is essential for the attention mechanism as it requires iterative training to find latent weights. However, the attention mechanism is hard to focus on the trend of the sequence efficiently, which results in the insufficient representation of the long-term behaviour of the node. As shown in Fig.3, The long-term trend line coloured purple can describe the global behaviour of the sequence, while the local trend line coloured green can be more flexible. And it is a fact that the overall trend of the node sequence is composed of a limited number of local trends. Thus, we design the Long-range temporal Strengthen module, shorten as $LT^2S$, to capture long-range dependencies of sequences, which plays an important role in the time series attention mechanism. It can be expressed as follows:
\begin{equation}
\begin{aligned}
X_{conv}=\ {\sigma(Conv}_{1\times1}\left(X_{low}\right))
\end{aligned}
\end{equation}
\begin{equation}
\begin{aligned}
LT^2S\left(X_{low}\right)={Concat}^{-1}(X_{low}^0,X_{conv},X_{low}^{T-1})
\end{aligned}
\end{equation}
where ${Concat}^{-1}(\cdot)$ represents the concatation of all tensors from the penultimate dimension, and $\sigma(\cdot)$ is the activation function.

The Long-term Temporal Strengthen module performs one-dimensional convolution on the time series of each node and constructs new time series features by splicing the traffic at the first and last time points, ensuring the dimensional consistency of input and output. With the multi-layer CST Blocks, the final block can hold temporal series features effectively. Finally, a TCN \cite{TCN} layer is used to predict the traffic data in the output layer.

\subsection{CST Block}
The modelling of spatiotemporal graph networks inherently has limitations, which are embodied in 1) the complex coupling of spatiotemporal graphs makes it almost impossible to take into account both the correlations between spatial and temporal dimensions; 2) the perception of high-dimensional spatiotemporal dependencies is limited and cannot be adapted to the changes brought by weak features; 3) capturing long-range dependencies lacks stability at large scales features, making it difficult to express trending features, i.e. the stable features in the changing environments.

To solve the above problems, the combined spatiotemporal block is proposed, as shown in Fig.2 (right). In this module, synchronous and asynchronous methods are combined effectively. It takes in the advantages of both while avoiding the information redundancy carried by synchronous and asynchronous. The CST Block takes the original feature X as input and outputs a tensor of the same shape. After the multi-layer stacking of the modules, the final output X equipped with a multi-long-term temporal strengthen effect is obtained. 
% The computational details contained in this module are described in \textbf{Algorithm 1}.

The perception of high-dimensional spatiotemporal dependence is the key to solving complex spatiotemporal coupling. The CST Block adopts a binary-way architecture to achieve a combination of synchronous and asynchronous. In the binary-path asynchronous architecture, in order to perceive spatial and temporal dependencies synchronously, the temporal information in $X_{High}$ is fused into the DVGL layer used to generate signal dissemination ability, and output $X_s$ as a representation of spatial dependencies. Besides, $X_{Low}$ is used for the dependency perception of the time dimension in terms of sensing time and space dependencies asynchronously, while the output $X_t$ is used as the representation of time correlations, and finally, the spatial dependency and time dependency are aggregated by Aggregation module, which generates the output tensor $X_{out}$ for the next block.

For a more powerful weight assignment to the sequence after the $L^2TS$ module, a self-attention mechanism is utilized to enhance the representation:

\begin{equation}
\begin{aligned}
E^{'}&=V_p\ \sigma(((x\ast_d\ \tau_1\ )\ W_p\ )(x\ast_d\ \tau_2\ )+b_p\ )
\end{aligned}
\end{equation}
\begin{equation}
\begin{aligned}
E&=Softmax(BN(E^{'}))
\end{aligned}
\end{equation}
where $V_{p} \in\mathbb{R}^{L \times L}$, $W_{p} \in\mathbb{R}^{N \times F}$ and $b_p \in \mathbb{R}^{L \times L}$. The attention score equipped with long-term temporal strengthen would be accumulated for the final prediction executed by TCN \cite{TCN}.

% \usepackage{multirow}
\begin{table*}[]\scriptsize
\centering
\caption{Experiment results of CDVGM and baselines on four benchmarks}
\begin{tabular}{ccccccccccccccc}
\hline
\multirow{2}{*}{Datasets} & \multirow{2}{*}{Pub./Year} & \multirow{2}{*}{Category} & \multicolumn{3}{c}{PEMSD3}                       & \multicolumn{3}{c}{PEMSD4}                       & \multicolumn{3}{c}{PEMSD7}                      & \multicolumn{3}{c}{PEMSD8}                      \\ \cline{4-15} 
                          &                            &                           & MAE            & RMSE           & MAPE           & MAE            & RMSE           & MAPE           & MAE            & RMSE           & MAPE          & MAE            & RMSE           & MAPE          \\ \hline
LSTM                      & IEEE 1997                  & Temporal                  & 18.04          & 29.33          & 25.17          & 26.24          & 39.86          & 21.31          & 27.05          & 40.08          & 16.43         & 20.89          & 31.94          & 18.35         \\
GCRN                      & ICLR 2017                  & Async                     & 18.20          & 28.80          & 21.93          & 23.64          & 35.56          & 18.83          & 26.72          & 38.62          & 14.80         & 19.21          & 28.62          & 15.76         \\
Gated-STGCN               & IJCAI 2018                 & Async                     & 18.12          & 29.07          & 23.55          & 24.94          & 37.71          & 20.07          & 26.88          & 39.35          & 15.62         & 20.05          & 30.28          & 17.06         \\
ASTGCN                    & AAAI 2019                  & Async                     & 16.99          & 28.17          & 18.85          & 21.37          & 33.23          & 15.78          & 24.37          & 36.84          & 11.04         & 18.05          & 26.85          & 12.09         \\
STSGCN\dag                   & AAAI 2020                  & Sync                      & 17.48          & 29.21          & 16.78          & 21.19          & 33.65          & 13.90          & 24.26          & 39.03          & 10.21         & 17.13          & 26.80          & 10.96         \\
DGCN                      & IEEE 2020                  & Async                     & 16.69          & 27.16          & 16.69          & 20.78          & 32.35          & 14.02          & 21.00          & 32.94          & 9.35          & 16.28          & 24.71          & 10.96         \\
STFGNN\dag                   & AAAI 2021                  & Sync                      & 16.77          & 28.34          & 16.30          & 19.83          & 31.88          & 13.02          & 22.07          & 35.80          & 9.21          & 16.64          & 26.22          & 10.60         \\
$S^2TAT$                     & IEEE 2022                  & Sync                      & 16.06          & 27.26          & 15.78          & 19.16          & 30.99          & 12.64          & 22.52          & 35.70          & 9.61          & 15.44          & 24.33          & 10.01         \\
FOGS\dag                     & IJCAI 2022                 & Learning                     & 15.06          & 24.25          & 14.11          & 19.35          & 31.33          & 12.71          & 20.62          & 33.96          & 8.58          & 14.92          & 24.09          & 9.42          \\ \hline
\textbf{CDVGM-CNN}        & \textbf{Ours}              & \textbf{Combine}          & \textbf{15.40} & \textbf{24.50} & \textbf{15.31} & \textbf{19.03} & \textbf{30.57} & \textbf{12.80} & \textbf{20.83} & \textbf{32.82} & \textbf{8.95} & \textbf{14.58} & \textbf{22.86} & \textbf{9.54} \\
\textbf{CDVGM-TCN}        & \textbf{Ours}              & \textbf{Combine}          & \textbf{15.75}      & \textbf{24.72}      & \textbf{15.50}      & \textbf{18.62} & \textbf{30.03} & \textbf{12.59} & \textbf{19.17}      & \textbf{30.66}      & \textbf{10.01}     & \textbf{14.64} & \textbf{22.76} & \textbf{9.22} \\ \hline
\end{tabular}
\end{table*}

\subsection{Aggregation and Fusion Layer}
The aggregation module is essential for asynchronous modelling approaches. With the global design, the fusion layer here is simple and the function is to transform the shape of the data for prediction executed by TCN \cite{TCN}. Therefore, two convolution and addition operations are adopted. Finally, the residual connection is used to protect the gradients, the LayerNorm is used for feature regularization and as mentioned above, the TCN is utilized for the target prediction.

% =======
% FIG. 04
% =======
%
\begin{figure*}[!htb]\scriptsize
\begin{center}
\includegraphics[width=1\textwidth,height=0.7\textwidth]{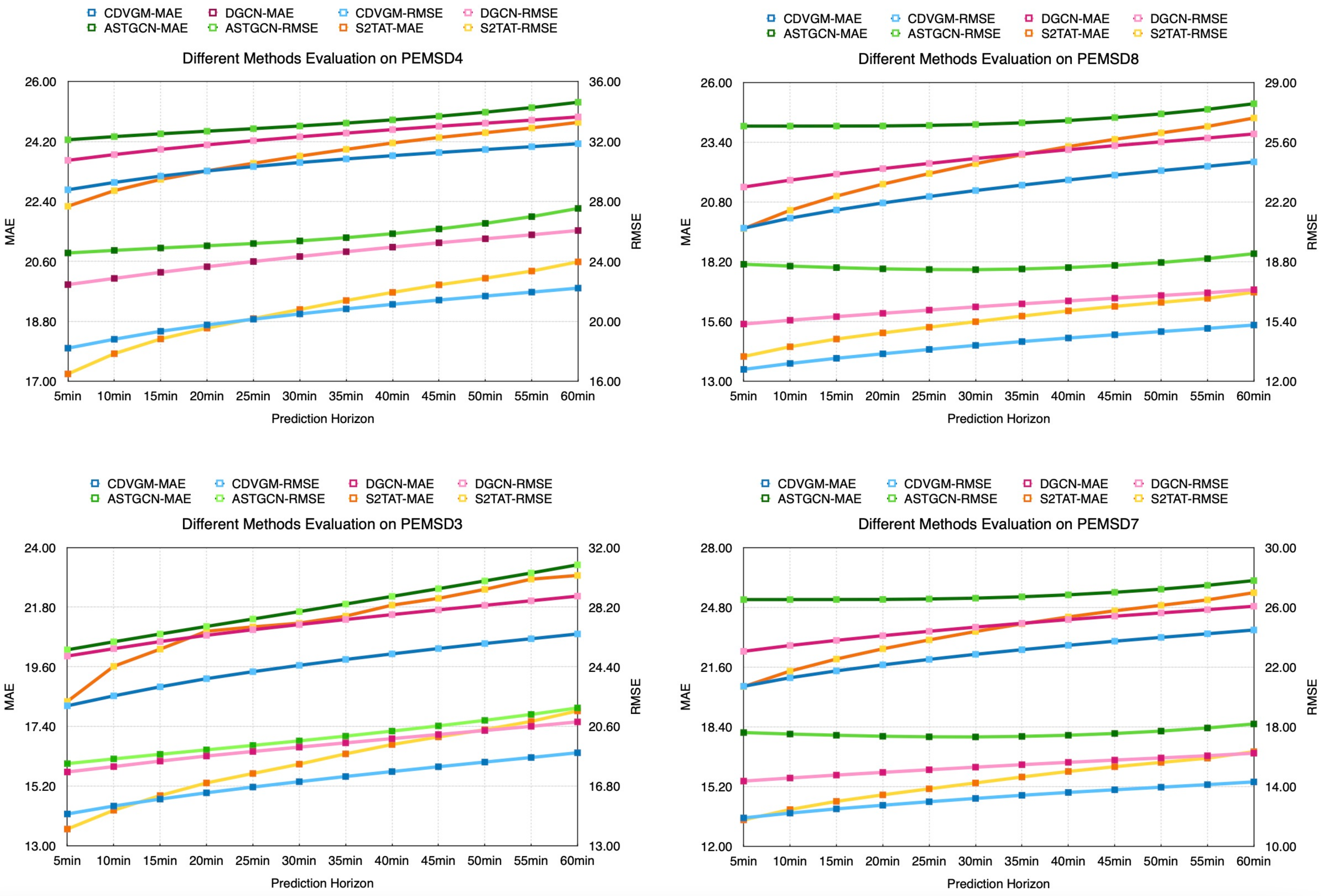}
\end{center}
\caption{The results of experiments on DGCN, ASTGCN, S2TAT and CDVGM on four benchmarks in the one-hour traffic flow prediction task. In each figure, the upper part is the curve of RMSE, the lower part is the curve of MAE, where the blue line represents CDVGM, which performs the best in all benchmarks.}
\end{figure*}

\section{Experiments}
\subsection{Datasets}
The experiments of CDVSTGM are conducted on four public benchmarks \cite{STSGCN}, PeMSD3, PeMSD4, PeMSD7 and PeMSD8 respectively. They are collected by Caltrans Performance Measurement System (PeMS) with 5min time slots and released by \cite{PeMS}. The details are listed in Table 2, which is collected in \cite{FOGS}, and $E/N$ is the ratio of the number of edges to the number of nodes. Except for the normalization in the CST block, we do not execute any standardization.

\begin{table}[!htbp]\scriptsize
\centering
\caption{Details of four benchmarks}
\begin{tabular}{c|c|c|c|c|c}
\hline
Dataset & \#Day & \#Nodes & \#Edges & \#Samples & \#E/N \\ \hline
PeMSD3  & 91          & 358     & 547     & 26208     & 1.53  \\
PeMSD4  & 59          & 307     & 340     & 16992     & 1.11  \\
PeMSD7  & 98          & 883     & 866     & 28224     & 0.98  \\
PeMSD8  & 62          & 170     & 295     & 17856     & 1.73  \\ \hline
\end{tabular}
\end{table}

\subsection{Baseline methods}
\begin{enumerate}
    \item [1)]LSTM (1997) \cite{LSTM}: Long-short Term Memory, a series prediction method, treats traffic forecasting as a sequential modelling task.
    \item [2)]GCRN (ICLR 2017) \cite{GCRN}: Graph Convolutional Recurrent Network, a generalized method of RNN in graph-based data asynchronously.
    \item [3)]Gated-STGCN (IJCAI 2018) \cite{STGCN}: Spatiotemporal Graph Convolutional Network, modelling traffic data from spatiotemporal views asynchronously.
    \item [4)]ASTGCN (AAAI 2019) \cite{ASTGCN}: A asynchronous method that is equipped with an attention mechanism for both spatial and temporal modelling.
    \item [5)]Gated-STSGCN (AAAI 2020) \cite{Gated_STGCN}: A method that models spatiotemporal correlations with local spatiotemporal graphs synchronously.
    \item [6)]DGCN (IEEE 2020) \cite{DGCN}: A framework that dynamically models Laplacian matrix for traffic forecasting asynchronously. 
    \item [7)]STFGNN (AAAI 2021) \cite{STFGNN}: Spatial-Temporal Fusion Graph Neural Network, modelling spatiotemporal dependencies with generated temporal graph synchronously. 
    \item [8)]$S^TAT$ (IEEE Trans. 2022) \cite{S2TAT}: Synchronous Spatio-Temporal Graph Transformer, known as its time-wise graph convolution and ability to capture non-local spatiotemporal relationships synchronously.
    \item [9)]FOGS (IJCAI 2022) \cite{FOGS}: a learning-based method to learn a spatial temporal correlation graph. It utilizes first-order gradients, rather than specific flows to train a model.
    
\end{enumerate}

\subsection{Experiment settings}
% We conduct the experiments on four widely used benchmarks to compare the proposed methods with existing approaches by the metrics of MAE, RMSE and MAPE, which are defined below:

The four datasets are split with a ratio of 6:2:2 and the forecasting horizon is one hour in the future, which contains twelve-time slots. We construct the CDVGM framework using Pytorch 1.8.1, and all models are executed on NVIDIA Geforce RTX 3090 GPU. We set the hyperparameter K as 4 in the ChebyNet. The batch size is fixed at 16 and the epoch is 40 except for $S^2TAT$ with 200. We train our model using the Adam optimizer with an initial learning rate of 0.001 and the LookAhead \cite{LookAhead} mechanism is utilized for effective optimization. Each method is conducted ten times and the result is the top. It is worth noting that the work marked with the \dag symbol has adopted the results in the paper because the code cannot be reproduced, which is not necessarily comparable to other methods.

To Evaluate the model, three metrics are defined which are the same as baselines, see Eq.(10), (11) and (12).

\begin{equation}
\begin{aligned}
    MAE\ =\ \frac{\sum_{i=1}^{T}\sum_{j=1}^{N}{|x_{ij}-{\hat{x}}_{ij}|}}{TN}
\end{aligned}
\end{equation}

\begin{equation}
\begin{aligned}
    RMSE\ =\ \sqrt{\frac{\sum_{i=1}^{T}\sum_{j=1}^{N}{(x_{ij}-{\hat{x}}_{ij})}^2}{TN}}
\end{aligned}
\end{equation}

\begin{equation}
MAPE\ =\ \frac{100\% \sum_{i=1}^{T}\sum_{j=1}^{N}\left|\frac{x_{ij}-{\hat{x}}_{ij}}{x_{ij}}\right|}{TN}
\end{equation}

The loss function is MSE, defined below, which considers the loss value of each time slot at the single node level.

\begin{equation}
\begin{aligned}
    MSE\ =\ \sqrt{\frac{\sum_{i=1}^{T}\sum_{j=1}^{N}{(x_{ij}-{\hat{x}}_{ij})}^2}{TN}}
\end{aligned}
\end{equation}

\begin{figure*}[!htb]
\begin{center}
\includegraphics[width=1\textwidth,height=0.46\textwidth]{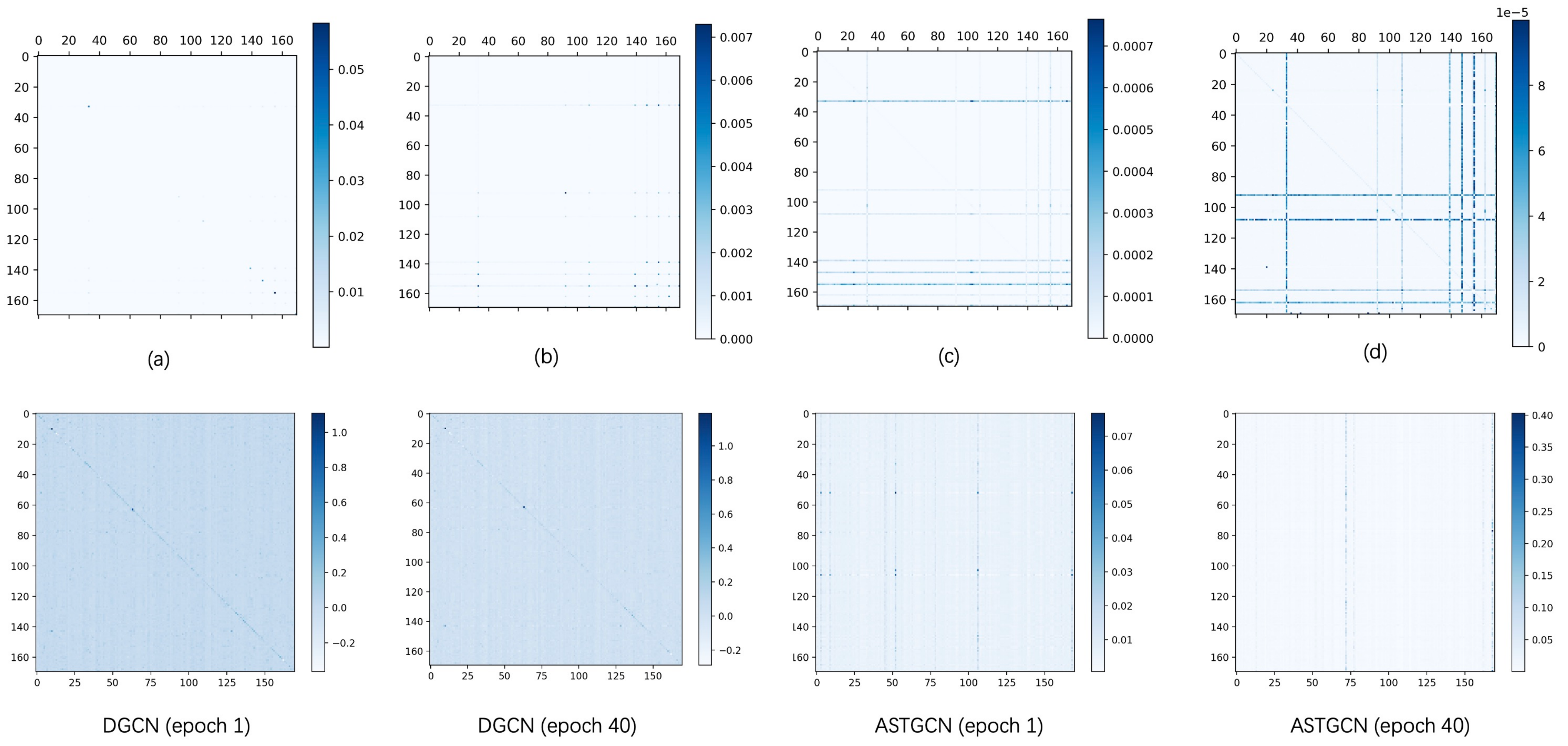}
\end{center}
\caption{The visualization on the Laplacian matrix generated by the DVGL module, DGCN and ASTGCN. Owing to the large gaps in the matrix value, large values are masked in (b), (c) and (d) for better visualization and (a) is the source matrix. Each of the above four figures from (a) to (d) represents the different horizons, indicating that the matrix can express the correlations of the nodes in different relation orders, i.e. the larger, the closer.}
\end{figure*}

\subsection{Experiment results}
The experiment results on four benchmarks are shown in Table I. From the table, the conclusion can be drawn that the method that only considers temporal relationships and performs the lowest accuracy among the predictions, while the methods that consider both spatial and temporal relationships are more powerful, which indicates that an excellent traffic prediction must consider both temporal and spatial dimensions. The gradient-based learning method FOGS is more capable than that but it pays attention to the road map which is prior defined.

Except for the temporal method and FOGS, other methods can be divided into three groups, namely asynchronous methods (A), synchronous methods (S) and combined methods (C). 

Among them, A-class methods are capable of temporal prediction, as it has a strong ability to decouple prediction task into structural-based distinguish and series-based forecasting. With the assumption of non-related orders on each node, the temporal module can independently execute prediction upon the node itself. 

Owing to the local receptive field and the way of pulling dependencies together, S-class methods are capable of node-oriented modelling, they pay more attention to limited horizons to reveal the complex coupling.

For C-class method, proposed in this work, namely the combination of A- and B-classes, is capable of the prediction task. It outperforms the baselines ASTGCN, STSGCN, DGCN, STFGNN, $S^2TAT$, and FOGS with a ratio of 16.8\%, 22.6\%, 6.9\%, 14.36\%, 14.12\%, 9.8\% respectively under the RMSE matric on PeMSD7 benchmark.

To evaluate the stability of the one-hour prediction task, the comparing experiments are conducted on four benchmarks and the results are visualized in Fig 5. The results indicate that even on the long-range horizon, the CDVGM also has the lowest prediction error, which implies an excellent long-term prediction ability. Among them, the $S^2TAT$ has the closest result with CDVGM, but it is not stable enough on different datasets while dealing with varying distributions of data. 

\section{Ablation Analysis}
The framework CDVGM proposed in this work has three components in total. Each of them has the ability to promote effectiveness while predicting. Therefore, ablation experiments are conducted to evaluate the performance further.

\subsection{How does DVGL perform?}
To explore the reason for the excellent result, the Laplacian matrix generated by the DVGL module has been visualized with random samples in the PeMSD8 dataset. As shown in Fig 6, the sparse matrix (from (a) to (d)) deliver the different levels of correlations among all nodes in the graph. Besides, the matrix is generated in each iteration with part of the dataset, which not only promotes the convergence speed but also leverages the related features effectively. Compared with other matrices generated by ASTGCN and DGCN (bottom), the matrix generated by CDVGM has more expressive information and highly considers the semantic relationship among different nodes.

The results demonstrate that the dynamics of Graph Laplacian in these three methods are in the different levels of expression, where DVGL is more expressive, which is closer to the real. Therefore, the matrix proposed in this work has the best expressive ability to conduct node selection and fusion. Compared with CDVGM, DGCN learns more noise and highly relies on the model initialization and ASTGCN lacks filtration on the signals of the inputs, which leads to bad prediction.

\subsection{Capability of $LT^2S$}
It's believed that long-range horizon is hard to conduct series prediction because of the uncertainty of dynamic changes. However, the trend can express the approximate direction of the series, which is useful for the prediction task. Regarding the $LT^2S$ mechanism proposed in this work, some experiments are conducted to explore its efficiency, as shown in Table 2 and 3. $LT^2S$ is divided into TS and TCN modules, to further explore the performance.

\begin{table}[!htbp]\scriptsize
\centering
\caption{Details of Ablations on $LT^2S$ (PEMSD4)}
\begin{tabular}{cccccc}
\hline
\multirow{2}{*}{DVGL} & \multirow{2}{*}{TS} & \multirow{2}{*}{TCN} & \multicolumn{3}{c}{PeMSD4}              \\ \cline{4-6} 
                     &                     &                      & MAE         & RMSE        & MAPE        \\ \hline
\checkmark                    &                     &                      & 19.05 (0.21) & 30.52 (0.22) & 12.86 (0.15) \\
\checkmark                    & \checkmark                   &                      & 18.88 (0.16) & 30.42 (0.17) & 12.73 (0.12) \\
\checkmark                    &                     & \checkmark                    & 19.25 (0.15) & 30.56 (0.13) & 13.11 (0.20) \\
\checkmark                    & \checkmark                   & \checkmark                    & 18.98 (0.23) & 30.44 (0.20) & 12.90 (0.23) \\ \hline
                     &                     &                      &             &             &            
\end{tabular}
\end{table}
\begin{table}[!htbp]\scriptsize
\centering
\caption{Details of Ablations on $LT^2S$ (PEMSD8)}
\begin{tabular}{cccccc}
\hline
\multirow{2}{*}{DVGL} & \multirow{2}{*}{TS} & \multirow{2}{*}{TCN} & \multicolumn{3}{c}{PeMSD8}              \\ \cline{4-6} 
                     &                     &                      & MAE         & RMSE        & MAPE        \\ \hline
\checkmark                    &                    &                     & 15.35 (0.28) & 23.62 (0.27) & 10.12 (0.25) \\
\checkmark                    & \checkmark                   &                     & 15.21 (0.32) & 23.45 (0.35) & 9.93 (0.36)  \\
\checkmark                    &                    & \checkmark                    & 14.97 (0.23) & 23.06 (0.22) & 9.49 (0.15)  \\
\checkmark                    & \checkmark                   & \checkmark                    & 14.81 (0.22) & 22.96 (0.23) & 9.39 (0.13)  \\ \hline
\end{tabular}
\end{table}

The above two experiments are conducted with a fixed seed (including PyTorch, Numpy and CUDNN) so that the initialization of the parameters is totally the same. Each experiment is conducted 10 times (with 10 random seeds) and evaluated by the average value of MAE, RMSE and MAPE (standard deviations are shown in parentheses). The module equipped with Temporal Strengthen ($TS$) can reduce 0.15 MAE on average on PeMSD8 and 0.14 MAE on average on PeMSD4, indicating that the $LT^2S$ module could efficiently improve the prediction accuracy and stability.

\section{Conclusion}
This paper proposes a novel method named CDVGM to capture spatiotemporal dependency in the way of both synchronous and asynchronous. Besides, a simple yet effective Combined Spatio-Temporal block (CST block), Dynamic Virtual Graph Laplacian (DVGL),  Long-term Temporal Strengthen ($LT^2S$) are proposed to explore latent and long-range correlations of spatial and temporal dimensions for traffic flow forecasting. By adopting these methods, the CDVGM has the powerful ability of adaptive graph representation. Extensive experiments demonstrate that it outperforms existing state-of-the-art methods on four benchmarks.

\section*{Acknowledgments}

I'm grateful to professors in Westlake University and Yunnan University, who taught me how to do scientific research with clearer targets and more effective methods no matter what kind of projects I'm working on, and thanks for the research platform provided by Westlake University, which allows me to be more efficient in scientific research and be closer to the masters for academic exchanges.
% \appendix

% \section{\LaTeX{} and Word Style Files}\label{stylefiles}

%
% 
% 
% The \LaTeX{} and Word style files are available on the IJCAI--19
% website, \url{http://www.ijcai19.org}.
% These style files implement the formatting instructions in this
% document.

% The \LaTeX{} files are {\tt ijcai19.sty} and {\tt ijcai19.tex}, and
% the Bib\TeX{} files are {\tt named.bst} and {\tt ijcai19.bib}. The
% \LaTeX{} style file is for version 2e of \LaTeX{}, and the Bib\TeX{}
% style file is for version 0.99c of Bib\TeX{} ({\em not} version
% 0.98i). The {\tt ijcai19.sty} style differs from the {\tt
% ijcai18.sty} file used for IJCAI--18.

% The Microsoft Word style file consists of a single file, {\tt
% ijcai19.doc}. This template differs from the one used for
% IJCAI--18.

% These Microsoft Word and \LaTeX{} files contain the source of the
% present document and may serve as a formatting sample.  

% Further information on using these styles for the preparation of
% papers for IJCAI--19 can be obtained by contacting {\tt
% pcchair@ijcai19.org}.

\bibliographystyle{named}
\bibliography{ijcai19}

\begin{thebibliography}{}

\bibitem[\protect\citeauthoryear{Azevedo \bgroup \em et al.\egroup
  }{2022}]{STref1}
Tiago Azevedo, Alexander Campbell, Rafael Romero-Garcia, Luca Passamonti,
  Richard~AI Bethlehem, Pietro Li{\`o}, and Nicola Toschi.
\newblock A deep graph neural network architecture for modelling
  spatio-temporal dynamics in resting-state functional mri data.
\newblock {\em Medical Image Analysis}, 79:102471, 2022.

\bibitem[\protect\citeauthoryear{Bai \bgroup \em et al.\egroup }{2018}]{TCN}
Shaojie Bai, J~Zico Kolter, and Vladlen Koltun.
\newblock An empirical evaluation of generic convolutional and recurrent
  networks for sequence modeling.
\newblock {\em arXiv preprint arXiv:1803.01271}, 2018.

\bibitem[\protect\citeauthoryear{Cao \bgroup \em et al.\egroup }{2022}]{STSSN}
Shuqin Cao, Libing Wu, Jia Wu, Dan Wu, and Qingan Li.
\newblock A spatio-temporal sequence-to-sequence network for traffic flow
  prediction.
\newblock {\em Information Sciences}, 610:185--203, 2022.

\bibitem[\protect\citeauthoryear{Chen \bgroup \em et al.\egroup }{2001}]{PeMS}
Chao Chen, Karl Petty, Alexander Skabardonis, Pravin Varaiya, and Zhanfeng Jia.
\newblock Freeway performance measurement system: mining loop detector data.
\newblock {\em Transportation Research Record}, 1748(1):96--102, 2001.

\bibitem[\protect\citeauthoryear{Chen \bgroup \em et al.\egroup }{2022}]{TM1}
Ling Chen, Wei Shao, Mingqi Lv, Weiqi Chen, Youdong Zhang, and Chenghu Yang.
\newblock Aargnn: An attentive attributed recurrent graph neural network for
  traffic flow prediction considering multiple dynamic factors.
\newblock {\em IEEE Transactions on Intelligent Transportation Systems}, 2022.

\bibitem[\protect\citeauthoryear{Defferrard \bgroup \em et al.\egroup
  }{2016}]{Chebynet}
Micha{\"e}l Defferrard, Xavier Bresson, and Pierre Vandergheynst.
\newblock Convolutional neural networks on graphs with fast localized spectral
  filtering.
\newblock {\em Advances in neural information processing systems}, 29, 2016.

\bibitem[\protect\citeauthoryear{Fang \bgroup \em et al.\egroup
  }{2021}]{STGODE}
Zheng Fang, Qingqing Long, Guojie Song, and Kunqing Xie.
\newblock Spatial-temporal graph ode networks for traffic flow forecasting.
\newblock In {\em Proceedings of the 27th ACM SIGKDD Conference on Knowledge
  Discovery \& Data Mining}, pages 364--373, 2021.

\bibitem[\protect\citeauthoryear{Gao \bgroup \em et al.\egroup }{2022}]{STref4}
Qiang Gao, Fan Zhou, Ting Zhong, Goce Trajcevski, Xin Yang, and Tianrui Li.
\newblock Contextual spatio-temporal graph representation learning for
  reinforced human mobility mining.
\newblock {\em Information Sciences}, 2022.

\bibitem[\protect\citeauthoryear{Guo \bgroup \em et al.\egroup
  }{2019a}]{ASTGCN}
Shengnan Guo, Youfang Lin, Ning Feng, Chao Song, and Huaiyu Wan.
\newblock Attention based spatial-temporal graph convolutional networks for
  traffic flow forecasting.
\newblock In {\em Proceedings of the AAAI conference on artificial
  intelligence}, volume~33, pages 922--929, 2019.

\bibitem[\protect\citeauthoryear{Guo \bgroup \em et al.\egroup
  }{2019b}]{ST-3DNet}
Shengnan Guo, Youfang Lin, Shijie Li, Zhaoming Chen, and Huaiyu Wan.
\newblock Deep spatial–temporal 3d convolutional neural networks for traffic
  data forecasting.
\newblock {\em IEEE Transactions on Intelligent Transportation Systems},
  20(10):3913--3926, 2019.

\bibitem[\protect\citeauthoryear{Guo \bgroup \em et al.\egroup }{2022}]{DGCN}
Kan Guo, Yongli Hu, Zhen Qian, Yanfeng Sun, Junbin Gao, and Baocai Yin.
\newblock Dynamic graph convolution network for traffic forecasting based on
  latent network of laplace matrix estimation.
\newblock {\em IEEE Transactions on Intelligent Transportation Systems},
  23(2):1009--1018, 2022.

\bibitem[\protect\citeauthoryear{Hamilton \bgroup \em et al.\egroup
  }{2017}]{GraphSage}
Will Hamilton, Zhitao Ying, and Jure Leskovec.
\newblock Inductive representation learning on large graphs.
\newblock {\em Advances in neural information processing systems}, 30, 2017.

\bibitem[\protect\citeauthoryear{Hochreiter and Schmidhuber}{1997}]{LSTM}
Sepp Hochreiter and J{\"u}rgen Schmidhuber.
\newblock Long short-term memory.
\newblock {\em Neural computation}, 9(8):1735--1780, 1997.

\bibitem[\protect\citeauthoryear{Huang \bgroup \em et al.\egroup
  }{2022}]{MSGAT}
Jing Huang, Kun Luo, Longbing Cao, Yuanqiao Wen, and Shuyuan Zhong.
\newblock Learning multiaspect traffic couplings by multirelational graph
  attention networks for traffic prediction.
\newblock {\em IEEE Transactions on Intelligent Transportation Systems}, pages
  1--15, 2022.

\bibitem[\protect\citeauthoryear{Jiang and Luo}{2022}]{IM4}
Weiwei Jiang and Jiayun Luo.
\newblock Graph neural network for traffic forecasting: A survey.
\newblock {\em Expert Systems with Applications}, page 117921, 2022.

\bibitem[\protect\citeauthoryear{Jin \bgroup \em et al.\egroup
  }{2022}]{AutoDSTSG}
Guangyin Jin, Fuxian Li, Jinlei Zhang, Mudan Wang, and Jincai Huang.
\newblock Automated dilated spatio-temporal synchronous graph modeling for
  traffic prediction.
\newblock {\em IEEE Transactions on Intelligent Transportation Systems}, 2022.

\bibitem[\protect\citeauthoryear{Junior \bgroup \em et al.\egroup
  }{2014}]{ARIMA}
Paulo~Rotela Junior, Fernando Luiz~Ri{\^e}ra Salomon, Edson
  de~Oliveira~Pamplona, et~al.
\newblock Arima: An applied time series forecasting model for the bovespa stock
  index.
\newblock {\em Applied Mathematics}, 5(21):3383, 2014.

\bibitem[\protect\citeauthoryear{Kipf and Welling}{2016}]{GCN}
Thomas~N Kipf and Max Welling.
\newblock Semi-supervised classification with graph convolutional networks.
\newblock {\em arXiv preprint arXiv:1609.02907}, 2016.

\bibitem[\protect\citeauthoryear{Li and Zhu}{2021}]{STFGNN}
Mengzhang Li and Zhanxing Zhu.
\newblock Spatial-temporal fusion graph neural networks for traffic flow
  forecasting.
\newblock In {\em Proceedings of the AAAI conference on artificial
  intelligence}, volume~35, pages 4189--4196, 2021.

\bibitem[\protect\citeauthoryear{Li \bgroup \em et al.\egroup }{2017}]{DCRNN}
Yaguang Li, Rose Yu, Cyrus Shahabi, and Yan Liu.
\newblock Diffusion convolutional recurrent neural network: Data-driven traffic
  forecasting.
\newblock {\em arXiv preprint arXiv:1707.01926}, 2017.

\bibitem[\protect\citeauthoryear{Lira \bgroup \em et al.\egroup
  }{2022}]{STref5}
Hernan Lira, Luis Mart{\'\i}, and Nayat Sanchez-Pi.
\newblock A graph neural network with spatio-temporal attention for
  multi-sources time series data: An application to frost forecast.
\newblock {\em Sensors}, 22(4):1486, 2022.

\bibitem[\protect\citeauthoryear{Liu \bgroup \em et al.\egroup }{2022}]{STref3}
Ryan~Wen Liu, Maohan Liang, Jiangtian Nie, Yanli Yuan, Zehui Xiong, Han Yu, and
  Nadra Guizani.
\newblock Stmgcn: Mobile edge computing-empowered vessel trajectory prediction
  using spatio-temporal multi-graph convolutional network.
\newblock {\em IEEE Transactions on Industrial Informatics}, 2022.

\bibitem[\protect\citeauthoryear{Luo \bgroup \em et al.\egroup }{2022}]{ESTNET}
Guiyang Luo, Hui Zhang, Quan Yuan, Jinglin Li, and Fei-Yue Wang.
\newblock Estnet: Embedded spatial-temporal network for modeling traffic flow
  dynamics.
\newblock {\em IEEE Transactions on Intelligent Transportation Systems}, 2022.

\bibitem[\protect\citeauthoryear{Medina-Salgado \bgroup \em et al.\egroup
  }{2022}]{IM3}
Boris Medina-Salgado, Eddy S{\'a}nchez-DelaCruz, Pilar Pozos-Parra, and
  Javier~E Sierra.
\newblock Urban traffic flow prediction techniques: A review.
\newblock {\em Sustainable Computing: Informatics and Systems}, page 100739,
  2022.

\bibitem[\protect\citeauthoryear{Nagy and Simon}{2018}]{IM2}
Attila~M Nagy and Vilmos Simon.
\newblock Survey on traffic prediction in smart cities.
\newblock {\em Pervasive and Mobile Computing}, 50:148--163, 2018.

\bibitem[\protect\citeauthoryear{Rao \bgroup \em et al.\egroup }{2022}]{FOGS}
Xuan Rao, Hao Wang, Liang Zhang, Jing Li, Shuo Shang, and Peng Han.
\newblock Fogs: First-order gradient supervision with learning-based graph for
  traffic flow forecasting.
\newblock In {\em Proceedings of International Joint Conference on Artificial
  Intelligence, IJCAI}. ijcai. org, 2022.

\bibitem[\protect\citeauthoryear{Song \bgroup \em et al.\egroup
  }{2020}]{STSGCN}
Chao Song, Youfang Lin, Shengnan Guo, and Huaiyu Wan.
\newblock Spatial-temporal synchronous graph convolutional networks: A new
  framework for spatial-temporal network data forecasting.
\newblock In {\em Proceedings of the AAAI Conference on Artificial
  Intelligence}, volume~34, pages 914--921, 2020.

\bibitem[\protect\citeauthoryear{Tan and Wang}{2019}]{GCRN}
Ke~Tan and DeLiang Wang.
\newblock Learning complex spectral mapping with gated convolutional recurrent
  networks for monaural speech enhancement.
\newblock {\em IEEE/ACM Transactions on Audio, Speech, and Language
  Processing}, 28:380--390, 2019.

\bibitem[\protect\citeauthoryear{Tang and Zeng}{2022}]{IM5}
Jinjun Tang and Jie Zeng.
\newblock Spatiotemporal gated graph attention network for urban traffic flow
  prediction based on license plate recognition data.
\newblock {\em Computer-Aided Civil and Infrastructure Engineering},
  37(1):3--23, 2022.

\bibitem[\protect\citeauthoryear{Tedjopurnomo \bgroup \em et al.\egroup
  }{2020}]{IM1}
David~Alexander Tedjopurnomo, Zhifeng Bao, Baihua Zheng, Farhana Choudhury, and
  Alex~Kai Qin.
\newblock A survey on modern deep neural network for traffic prediction:
  Trends, methods and challenges.
\newblock {\em IEEE Transactions on Knowledge and Data Engineering}, 2020.

\bibitem[\protect\citeauthoryear{Velickovic \bgroup \em et al.\egroup
  }{2017}]{GAT}
Petar Velickovic, Guillem Cucurull, Arantxa Casanova, Adriana Romero, Pietro
  Lio, and Yoshua Bengio.
\newblock Graph attention networks.
\newblock {\em stat}, 1050:20, 2017.

\bibitem[\protect\citeauthoryear{Wang \bgroup \em et al.\egroup }{2022a}]{TM2}
Hanqiu Wang, Rongqing Zhang, Xiang Cheng, and Liuqing Yang.
\newblock Hierarchical traffic flow prediction based on spatial-temporal graph
  convolutional network.
\newblock {\em IEEE Transactions on Intelligent Transportation Systems}, 2022.

\bibitem[\protect\citeauthoryear{Wang \bgroup \em et al.\egroup
  }{2022b}]{S2TAT}
Tian Wang, Jiahui Chen, Jinhu L{\"u}, Kexin Liu, Aichun Zhu, Hichem Snoussi,
  and Baochang Zhang.
\newblock Synchronous spatiotemporal graph transformer: A new framework for
  traffic data prediction.
\newblock {\em IEEE Transactions on Neural Networks and Learning Systems},
  2022.

\bibitem[\protect\citeauthoryear{Wang \bgroup \em et al.\egroup
  }{2022c}]{STref2}
Wenhan Wang, Youyong Kong, Zhenghua Hou, Chunfeng Yang, and Yonggui Yuan.
\newblock Spatio-temporal attention graph convolution network for functional
  connectome classification.
\newblock In {\em ICASSP 2022-2022 IEEE International Conference on Acoustics,
  Speech and Signal Processing (ICASSP)}, pages 1486--1490. IEEE, 2022.

\bibitem[\protect\citeauthoryear{Wu \bgroup \em et al.\egroup }{2019}]{SGC}
Felix Wu, Amauri Souza, Tianyi Zhang, Christopher Fifty, Tao Yu, and Kilian
  Weinberger.
\newblock Simplifying graph convolutional networks.
\newblock In {\em International conference on machine learning}, pages
  6861--6871. PMLR, 2019.

\bibitem[\protect\citeauthoryear{Yu \bgroup \em et al.\egroup }{2017a}]{STGCN}
Bing Yu, Haoteng Yin, and Zhanxing Zhu.
\newblock Spatio-temporal graph convolutional networks: A deep learning
  framework for traffic forecasting.
\newblock {\em arXiv preprint arXiv:1709.04875}, 2017.

\bibitem[\protect\citeauthoryear{Yu \bgroup \em et al.\egroup
  }{2017b}]{Gated_STGCN}
Bing Yu, Haoteng Yin, and Zhanxing Zhu.
\newblock Spatio-temporal graph convolutional networks: A deep learning
  framework for traffic forecasting.
\newblock {\em arXiv preprint arXiv:1709.04875}, 2017.

\bibitem[\protect\citeauthoryear{Zhang \bgroup \em et al.\egroup
  }{2017}]{ST-ResNet}
Junbo Zhang, Yu~Zheng, and Dekang Qi.
\newblock Deep spatio-temporal residual networks for citywide crowd flows
  prediction.
\newblock In {\em Thirty-first AAAI conference on artificial intelligence},
  2017.

\bibitem[\protect\citeauthoryear{Zhang \bgroup \em et al.\egroup
  }{2019}]{LookAhead}
Michael Zhang, James Lucas, Jimmy Ba, and Geoffrey~E Hinton.
\newblock Lookahead optimizer: k steps forward, 1 step back.
\newblock {\em Advances in neural information processing systems}, 32, 2019.

\bibitem[\protect\citeauthoryear{Zhang \bgroup \em et al.\egroup
  }{2022}]{DynamicGraph}
Wenyu Zhang, Kun Zhu, Shuai Zhang, Qian Chen, and Jiyuan Xu.
\newblock Dynamic graph convolutional networks based on spatiotemporal data
  embedding for traffic flow forecasting.
\newblock {\em Knowledge-Based Systems}, page 109028, 2022.

\end{thebibliography}

\end{document}